\documentclass{article}
\usepackage{ijcai25}

\usepackage{times}
\usepackage{soul}
\usepackage{url}
\usepackage[hidelinks]{hyperref}
\usepackage[utf8]{inputenc}
\usepackage[small]{caption}
\usepackage{graphicx}
\usepackage{amsmath}
\usepackage{amsthm}
\usepackage{booktabs}
\usepackage{algorithm}
\usepackage{algorithmic}
\usepackage[switch]{lineno}
\usepackage{multirow}
\usepackage{amsfonts}
\usepackage{bm}
\usepackage {makecell}

\title{FissionVAE: Federated Non-IID Image Generation with Latent Space and Decoder Decomposition}

\author{
Chen Hu$^1$
\and
Hanchi Ren$^1$\and
Jingjing Deng$^2$\and
Xianghua Xie$^1$\And
Xiaoke Ma$^3$\\
\affiliations
$^1$Swansea University \ $^2$Durham University\ $^3$Xi'Dian University\\
\emails
\{chen.hu, hanchi.ren, x.xie\}@swansea.ac.uk,
jingjing.deng@durham.ac.uk
}

\begin{document}

\maketitle

\begin{abstract}
Federated learning is a machine learning paradigm that enables decentralized clients to collaboratively learn a shared model while keeping all the training data local. While considerable research has focused on federated image generation, particularly Generative Adversarial Networks, Variational Autoencoders have received less attention. In this paper, we address the challenges of non-IID (independently and identically distributed) data environments featuring multiple groups of images of different types. Non-IID data distributions can lead to difficulties in maintaining a consistent latent space and can also result in local generators with disparate texture features being blended during aggregation. We thereby introduce FissionVAE that decouples the latent space and constructs decoder branches tailored to individual client groups. This method allows for customized learning that aligns with the unique data distributions of each group. Additionally, we incorporate hierarchical VAEs and demonstrate the use of heterogeneous decoder architectures within FissionVAE. We also explore strategies for setting the latent prior distributions to enhance the decoupling process. To evaluate our approach, we assemble two composite datasets: the first combines MNIST and FashionMNIST; the second comprises RGB datasets of cartoon and human faces, wild animals, marine vessels, and remote sensing images. Our experiments demonstrate that FissionVAE greatly improves generation quality on these datasets compared to baseline federated VAE models.
\end{abstract}

\section{Introduction}

Generative models have attracted increasing attention in recent years due to their impressive ability to generate new data across various modalities, including images \cite{ddpm}, texts \cite{llama}, and audios \cite{audio}. As these models, like other deep learning systems, require substantial amounts of data, concerns regarding data privacy have elevated among regulatory authorities and the public. Unlike the traditional centralized learning paradigm, which collects all data on a single computer system for training, federated learning allows private data to remain on the owner’s device. In this paradigm, local devices train models independently, and a central server aggregates these models without accessing the individual data directly. Although this distributed approach enhances privacy protection, it also introduces unique challenges not encountered in centralized systems. Since data remains distributed across various client devices, the training samples are not guaranteed to be identically distributed. This can lead to inconsistencies in learning objectives among clients, resulting in degraded performance when these models are aggregated on the server. 

In the context of FL with non-IID data, generative models such as Generative Adversarial Networks (GANs) \cite{gan} and Variational Autoencoders (VAEs) \cite{vae-origin} face additional challenges. These models involve sampling from a latent distribution, and the generator or decoder trained on client devices may develop differing interpretations of the same latent space. This discrepancy can lead to difficulties in maintaining a consistent and unified latent space, resulting in ambiguous latent representations. A further challenge arises from the role of the generator or decoder, which are tasked with mapping latent inputs to the sample space by synthesizing the shape, texture, and colors of images. Aggregating generative models trained on non-IID image data can produce artifacts that appear as a blend of disparate image types, because generators trained on non-IID local data capture the characteristics of varied visual features. Specifically for GANs, another problem arises from local discriminators, which may provide conflicting feedback that hinders model convergence. With the limited data available in FL settings, discriminators can quickly overfit to the training samples \cite{ada}. If an updated generator from the server produces images of classes not present in a client's local dataset, the local discriminator might incorrectly label well-generated images as fake, simply because they do not match the local data distribution. This mislabeling can significantly impede the generator's ability to synthesize realistic images.

Existing research on generative models for non-IID data in federated learning (FL) has primarily focused on GANs. MDGAN \cite{mdgan} proposes exchanging local discriminators among clients during training. This strategy allows discriminators to access a broader spectrum of local data, thereby avoiding biased feedback to the generator. The authors of \cite{forgivingan} uses the local discriminator that gives the highest score to a generated sample to update the global generator, promoting the idea that local discriminators should only judge samples from familiar distributions. In \cite{groupGAN}, the authors aggregate generators at the group level for client groups sharing similar data distributions before performing a global aggregation, then the global generator is aggregated similar to \cite{forgivingan}. Both \cite{forgivingan} and \cite{groupGAN} involve sending synthesized samples back to local clients, which could potentially increase the risk of compromising client data privacy.

Studies employing VAEs solely for image generation purposes are less common. The works in \cite{vae-aid} and \cite{vae-kd} utilize VAEs to produce synthetic images that assist in training global classifiers. In \cite{vae-aid}, the global decoder generates minority samples for local classifiers by sampling from class means with added noise. The approach in \cite{vae-kd} treats converged local decoders as teacher models and uses knowledge distillation to train a global generator on the server side without further local updates. While this decoder can produce useful samples for classification tasks, it risks overfitting to the potentially flawed output from local decoders and lacks generative diversity, which is crucial for high-quality image generation. Recent studies \cite{image-collapse} \cite{llm-collapse} have shown that generative models trained on generated samples instead of real data are prone to collapsing.

In response to the challenges posed by non-IID data in federated image generation, we introduce a model named FissionVAE. This model is specifically tailored to environments featuring multiple groups of images of different types. To mitigate the problem of mixed latent space interpretation, FissionVAE decomposes the latent space into distinctive priors, hence adapting to the diverse data distributions across different image types. We further refine this approach by investigating strategies for encoding the prior Gaussians. Additionally, to prevent the blending of unrelated visual features in the generated outputs, FissionVAE employs specialized decoder branches for each client group. This method not only accommodates the unique characteristics of each data subset but also enhances the model's generative capabilities in highly heterogeneous environments. The primary contributions of our research are detailed as follows:

1. We introduce FissionVAE for federated non-IID image generation. In FissionVAE, we decompose the latent space according to the distinct data distributions of client groups. This approach ensures that each client's data are mapped to its corresponding latent distribution without the adverse effects of averaging dissimilar distributions during aggregation. Moreover, by implementing separate decoder branches for different groups of data, FissionVAE allows for specialized generation tailored to different image types, which is crucial for preserving the distinct visual features of different image types during the generative process.

2. We explore various strategies for encoding Gaussian priors to enhance the effectiveness of latent space decomposition. We further extends FissionVAE by introducing the hierarchical inference architecture. We demonstrate that with the decomposed decoder branches, it is feasible to employ heterogeneous decoder architectures in FissionVAE, allowing for more flexible model deployment on clients.

3. We validate FissionVAE with extensive experiments on two composite datasets combining MNIST with FashionMNIST, and a more diverse set comprising cartoon and human faces, animals, marine vessels, and remote sensing images. Our results demonstrate improvements in generation quality over the existing baseline federated VAE.

The remainder of the paper is organized as follows: In Section 2, we describe the baseline FedVAE model and the FissionVAE variants we propose. Section 3 presents the experimental setup, including the configuration details and an analysis of the results. Finally, we conclude the paper in Section 4 with a summary of our findings and a discussion on potential future directions.

\begin{figure*}[h]
\centering
\includegraphics[width=0.85\linewidth]{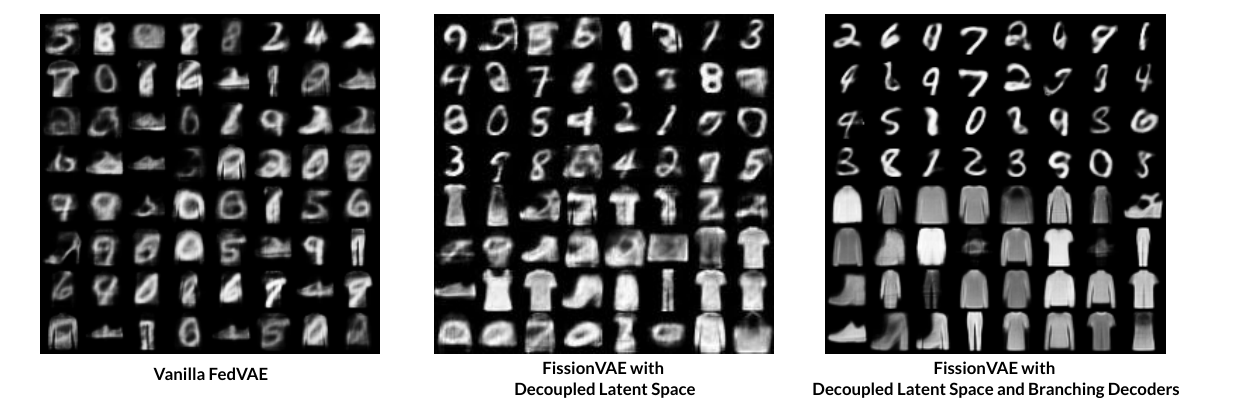}
\caption{Qualitative results of the baseline FedVAE and proposed FissionVAEs. As we further decoupling the latent space and decoders in the federated environment, the quality of generated images is improved.}
\label{fig: fedvae_all}
\end{figure*}

\section{Investigating Strategies for Non-IID Image Generation with VAEs}

In this section, we describe our methodology for exploring VAE configurations tailored for generating images under non-IID conditions in a federated learning framework. For background on FL and VAEs, please refer to the supplementary material. We specifically address scenarios where clients are categorized based on distinct data distributions. For illustrative purposes, we consider the case where some clients exclusively possess hand-written digit images from the MNIST dataset, while others maintain only clothing images from the FashionMNIST dataset. We follow to the standard federated learning framework, wherein a central server is tasked with aggregating updates from the clients and subsequently distributing the updated model back to them. FedAvg \cite{fedavg} is employed for server-side aggregation. Each client retains a subset of data representative of its respective group and conducts local training independently. A more practical scenario with RGB images and a larger number of client groups is explored and discussed in the experiments section (Section 3).

\subsection{FedVAE}

\begin{figure}[h]
\centering
\includegraphics[width=0.95\linewidth]{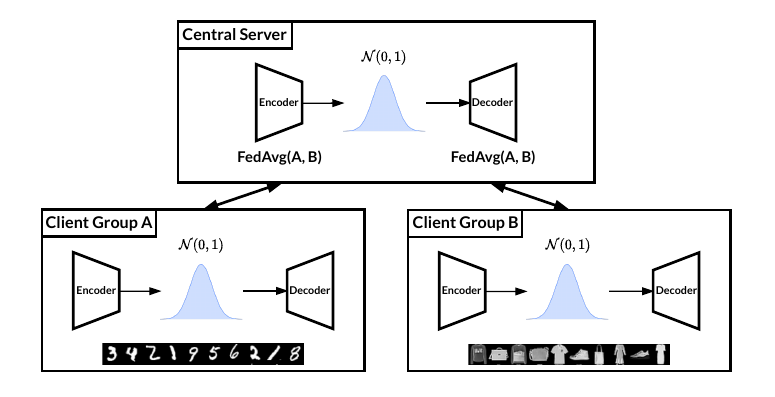}
\caption{An illustration of baseline FedVAE. The encoder and the decoder of the VAE are aggregated through FedAvg regardless of their client groups.}
\label{fig: vanilla}
\end{figure}

A straightforward strategy for implementing  VAEs in federated learning is using a unified encoder-decoder architecture. In this configuration, all clients share a common latent space (often predefined as the normal distribution $\mathcal{N}(0, 1)$) and the central server indiscriminately aggregates client models at the end of each training round. Fig. \ref{fig: vanilla} illustrates this baseline training scheme.

Despite the simplicity of this strategy, it present significant challenges in the non-IID scenario. Specifically, employing a single prior distribution for the latent space does not account for the distinct data distributions across different clients. Encoders from different client groups may map their uniquely distributed data into the same region of the latent space. Consequently, client decoders might interpret this shared latent space differently, leading to inconsistencies or even conflicts among client models during aggregation at the server. Figure \ref{fig: fedvae_all} shows randomly generated samples produced after training the federated Vanilla VAE on the combined dataset of MNIST and FashionMNIST. These samples clearly exhibit artifacts that appear to blend features of handwritten digits with clothing items, indicating the aggregation conflicts inherent in this method.

\subsection{FissionVAE with Latent Space Decoupling}
\begin{figure}[ht]
\centering
\includegraphics[width=0.95\linewidth]{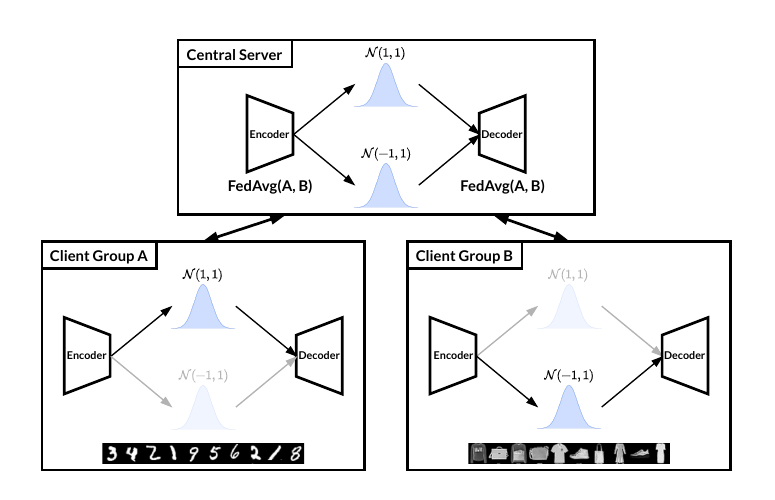}
\caption{An illustration of FissionVAE with Latent Space Decoupling. The latent variables are forced to follow their respective group prior distributions. The model is aggregated the same way as the baseline FedVAE.}
\label{fig: loom}
\end{figure}

To address the conflicting latent space issue identified above, we propose decomposing the latent space according to different data groups, while maintaining a unified architecture for the encoder and decoder. This approach corresponds to the architecture shown in Fig. \ref{fig: loom}.

\begin{figure*}[ht!]
\centering
\includegraphics[width=1.0\linewidth]{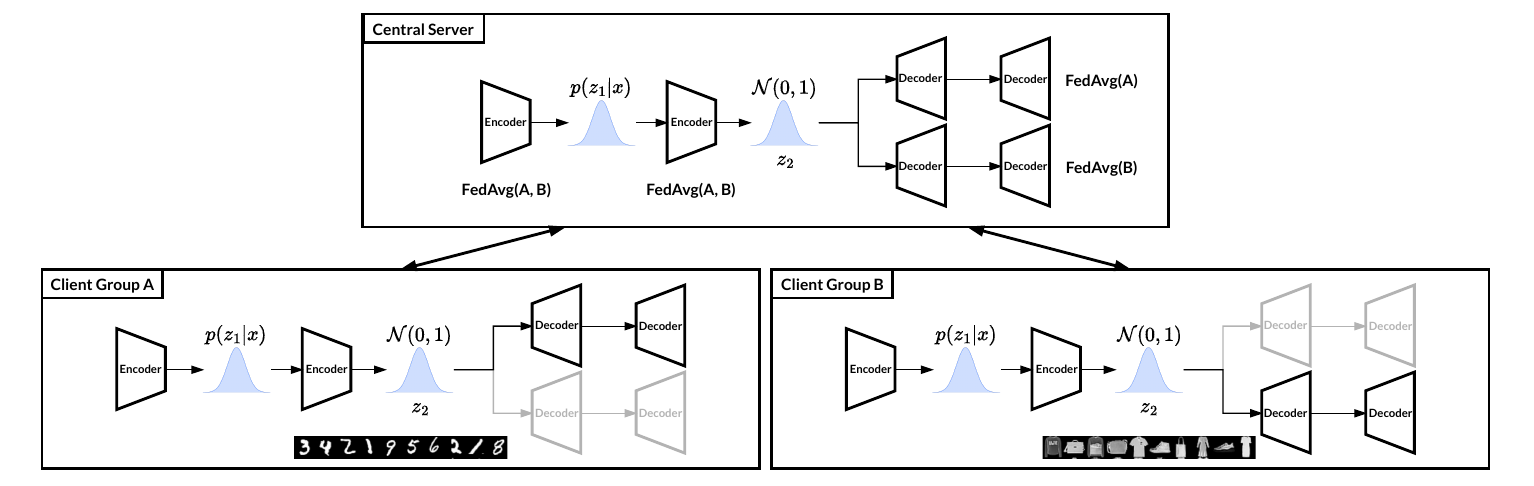}
\caption{An illustration of Hierarchical FissionVAE. This FissionVAE architecture extends to allow two levels of latent variables. The latent variable $z_1$ can be either learned or predefined. As input from different groups has been separated by $z_1$, the latent variable $z_2$ is set to follow the standard normal distribution.}
\label{fig: two_stage}
\end{figure*}

When decoupling the latent space, the encoder maps the input data to different distributions based on the client's group. For instance, MNIST client may map to $\mathcal{N}(-1, 1)$ and FashionMNIST clients to $\mathcal{N}(-1, 1)$. The KL divergence in the ELBO for this model is given by:

\begin{equation}
    D_{\text{KL}} (\mathcal{N}(\mu_q, \sigma_q||\mathcal{N}(\pm 1, 1)) = \frac{1}{2} \Sigma_{i=1}^{k} [\sigma_i + \mu_i^2 \mp 2\mu_i-\log \sigma_i]
\end{equation}

\noindent Here, $\mu_q$ and $\sigma_q$ represent the encoder's estimates for the parameters of the latent code's distribution, and $k$ is the dimension of the latent code.

Figure \ref{fig: fedvae_all} shows randomly generated amples produced after training the FissionVAE with latent space decoupling on the Mixed MNIST dataset. While the quality of reconstructed images are improved compared to the baseline FedVAE, the generated images still exhibit a mixture of handwritten digits and clothing items, even when explicitly sampling from their respective latent distributions. This suggests that while decomposing latent encoding helps improving reconstructions, the unified decoder still blends features due to the aggregation of model weights from diverse visual domains. This observation motivates the architecture described in the next section, where the decoder is also split based on client groups.

\subsection{FissionVAE with Group-specific Decoder Branches}

\begin{figure}[ht]
\centering
\includegraphics[width=1.0\linewidth]{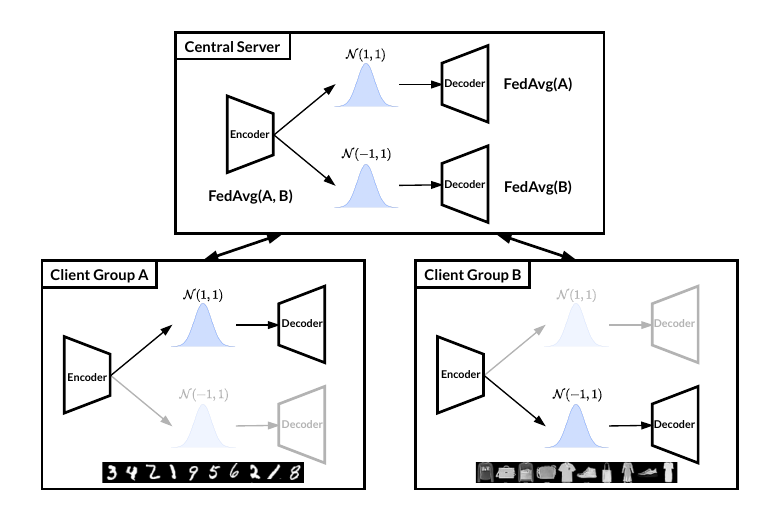}
\caption{An illustration of FissionVAE with Decoder Branch Decoupling. This FissionVAE creates decoders specific to client groups and enforces constraints for latent variable priors. The encoder is aggregated across groups while the group-specific decoder is only aggregated from local models within the corresponding group.}
\label{fig: two_way}
\end{figure}

\noindent \textbf{Non-Hierarchical FissionVAE} Building on the concept introduced by FissionVAE with latent space decoupling, we further refines non-IID data generation by incorporating decoder branches specific to each data group while maintaining a unified encoder. This design allows the central server to aggregate the encoder updates agnostically of the client groups, whereas decoder branches are aggregated specifically according to their corresponding groups. In addition, this approach also offers flexibility in the choice of the prior latent distribution $p(z)$ for each group to exert more explicit control over the data generation through the decoder. Figure \ref{fig: two_way} illustrates this branching architecture.

Figure \ref{fig: fedvae_all} also includes randomly generated samples produced after training the FissionVAE with decoder branches. The results indicate a significant reduction in the blending feature issue in previously discussed VAE architectures.

\noindent \textbf{Hierachical FissionVAE} Next, we show that the branching architecture can be enhanced by integrating hierarchical inference \cite{hvae-0} \cite{hvae-1} to the federated learning framework, which enables the use of deeper network structures to capture more complex data distributions. Fig \ref{fig: two_stage} depicts the FissionVAE with two levels of hierarchical inference. In this architecture, the first encoder module estimates $q(z_1|x)$ from the input data, then the second encoder module estimates $q(z_2|z_1)$ based on the first level latent code. The decoder reverses the encoding process, which estimates $p(z_1|z_2)$ based on $z_2$ to reconstruct $z_1$, and subsequently reconstructs the original input $x$ by estimating $p(x|z_1)$. 

Following the convention in hierarchical VAEs, we assume conditional independence among the latent codes.Then the ELBO for this hierarchical VAE is expressed as (refer to supplementary material for derivation), 

\begin{multline}
    \text{ELBO}_{H} = \mathbb{E}_{q_{\phi}(z_1|x)} [\log p_{\theta}(x|z_1)] \\ 
    - \mathbb{E}_{q_{\phi}(z_{1}|x)} [D_{\text{KL}} (q_{\phi}(z_2|z_1) || p(z_2))] \\ 
    - \mathbb{E}_{q_{\phi}(z_2|z_1)} [D_{\text{KL}}(q_{\phi}(z_1|x)||p_{\theta}(z_1|z_2)]
\end{multline}

\noindent In the equation above, the first term is the reconstruction term as it is the expectation of the log-likelihood for the input samples under the distribution estimated from the encoded $z_1$, the second term is the prior matching term which is enforcing the encoded $z_2$ to conform the prior distribution $z_2 \sim \mathcal{N}(0, 1)$, and the last term is the consistency term which requires $z_1$ from either the encoder or the decoder to be consistence. In practice, we find that adding the reconstruction loss from $z_2$ to $x$ is also crucial for generating meaningful samples. Optionally, perceptual losses such as the VGG loss \cite{vgg-loss} or the structural similarity index measure (SSIM) \cite{ssim} loss can be used to promote the fidelity of reconstructed images. However, no significant improvement is observed in our experiments. Therefore no perceptual loss is included in our implementation. The final loss function for the hierarchical and branching FissionVAE then becomes,

\begin{multline}
    \mathcal{L} = \mathbb{E}_{q_{\phi}(z_{1}|x)} [D_{\text{KL}} (q_{\phi}(z_1|z_x) || p(z_1))] \\ 
    - \mathbb{E}_{q_{\phi}(z_2|z_1)} [\log p_{\theta}(x|z_1, z_2)] - \text{ELBO}_{H}
\end{multline}

\noindent Here we minimize the KL divergence for $z_1$ only when the prior distribution for $z_1$ is explicitly defined, otherwise the model learns the latent distribution by itself.

\begin{table*}[t]
\centering
\resizebox{\textwidth}{!}{
\begin{tabular}{ccccccccccccc}
\hline
\multicolumn{1}{c|}{\multirow{3}{*}{Model}} & \multicolumn{6}{c|}{Mixed MNIST} & \multicolumn{6}{c}{CHARM} \\
\cline{2-13} 
\multicolumn{1}{c|}{} & \multicolumn{3}{c|}{Federated} & \multicolumn{3}{c|}{Centralized} & \multicolumn{3}{c|}{Federated} & \multicolumn{3}{c}{Centralized} \\
\cline{2-13} 
\multicolumn{1}{c|}{} & \multicolumn{1}{c|}{FID $\downarrow$} & \multicolumn{1}{c|}{IS $\uparrow$} & \multicolumn{1}{c|}{NLL $\downarrow$} & \multicolumn{1}{c|}{FID $\downarrow$} & \multicolumn{1}{c|}{IS $\uparrow$} & \multicolumn{1}{c|}{NLL $\downarrow$} & \multicolumn{1}{c|}{FID $\downarrow$} & \multicolumn{1}{c|}{IS $\uparrow$} & \multicolumn{1}{c|}{NLL $\downarrow$} & \multicolumn{1}{c|}{FID $\downarrow$} & \multicolumn{1}{c|}{IS $\uparrow$} & NLL $\downarrow$ \\
\hline
FedGAN & 118.52 & 2.39 & - & 91.08 & \underline{3.18} & - & - & - & - & - & - & - \\
FedVAE & 117.03 & 2.29 & \underline{0.23} & 40.59 & \textbf{3.62} & \textbf{0.18} & 167.18 & 1.57 & 40.80 & 89.26 & 2.57 & 46.99 \\
FissionVAE+L & 64.99 & 2.83 & \textbf{0.22} & 39.27 & 3.03 & \underline{0.18} & 155.81 & 1.73 & 43.49 & 86.19 & 2.53 & 51.45 \\
FissionVAE+D & \textbf{40.78} & 3.01 & 0.26 & 34.76 & 3.05 & 0.25 & 120.39 & 2.16 & \underline{33.07} & \underline{63.25} & \textbf{2.95} & \underline{36.76} \\
FissionVAE+L+D & \underline{42.11} & \textbf{3.04} & 0.25 & \underline{34.39} & 3.08 & 0.20 & \underline{109.10} & \underline{2.27} & 33.29 & \textbf{50.30} & \underline{2.89} & 40.14 \\
FissionVAE+H+L+D & 47.72 & \underline{2.98} & 0.30 & \textbf{28.82} & 3.16 & 0.24 & \textbf{107.69} & \textbf{2.32} & \textbf{27.46} & 74.59 & 2.58 & \textbf{27.09} \\
 \hline
\end{tabular}
}
\caption{Evaluation of proposed FissionVAEs on the Mixed MNIST and CHARM dataset. +L is for decoupled latent space. +D is for branching decoders. +H is for the hierarchical architecture. Best results in are in \textbf{bold}. Second best results are \underline{underlined}. $\uparrow$ denotes the higher the better, while $\downarrow$ means the lower the better.}
\label{overall}
\end{table*}

The proposed hierarchical FissionVAE also allows heterogeneous decoder architectures for each client groups, as each decoder branch is trained and aggregated independently. This flexibility is particularly advantageous in federated learning environments, where clients often possess varying computational resources. Client groups with more resources can implement deeper and more complex network structures, while groups with limited computational capacity can utilize lighter models.

\section{Experiments}

\subsection{Datasets and Evaluation Metrics}

We evaluated the proposed federated VAEs using two composite datasets. Mixed MNIST combines MNIST \cite{mnist} and FashionMNIST \cite{fashion}, dividing samples into two client groups (one per dataset) with 10 clients each. Training samples were evenly distributed within each group, and the default test sets served as evaluation benchmarks. An equal number of images were generated using the global model for comparison.

CHARM is a more diverse dataset combining five domains: Cartoon faces \cite{anime}, Human faces \cite{celeba-hq}, Animals \cite{animals}, Remote sensing images \cite{earth}, and Marine vessels \cite{boats}, using preprocessed square images from Meta-Album for AwA2 and MARVEL. Images were resized to $32 \times 32$, and each domain was represented by 20 clients, with 20,000 images for training and 5,000 for evaluation. As with Mixed MNIST, the global model generated evaluation samples.

For Mixed MNIST, encoders and decoders used Multi-Layer Perceptrons (MLPs). On CHARM, encoders $q(z_1|x)$ and decoders $p(x|z_1)$ were convolutional, while $q(z_2|z_1)$ and $p(z_1|z_2)$ used MLPs. Client participation followed a Bernoulli distribution: $B(0.5)$ for Mixed MNIST and $B(0.25)$ for CHARM. Hyperparameters included learning rates of $1 \times 10^{-3}$ (Mixed MNIST) and $1 \times 10^{-4}$ (CHARM), with 70 and 500 training rounds, respectively. Clients performed 5 local epochs per round with a batch size of 32. Centralized settings used 70 epochs for Mixed MNIST and 250 for CHARM.

Evaluation metrics included Fréchet Inception Distance \cite{fid} and Inception Score \cite{inception-score} for generation quality, and the negative log-likelihood (NLL) of the ELBO for reconstruction performance. IS was computed using an ImageNet-pretrained Inception model \cite{inception}.

\begin{figure*}[h!]
\centering
\includegraphics[width=1.0\linewidth]{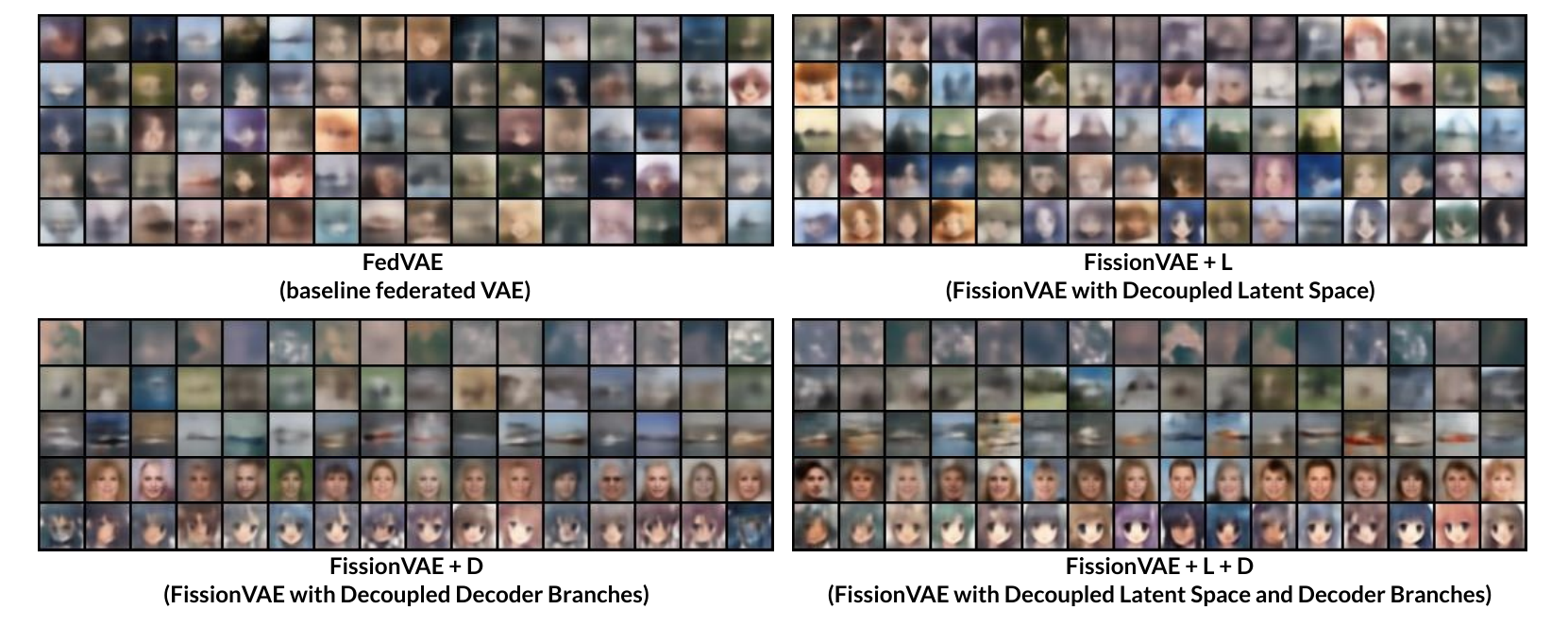}
\caption{Qualitative results of image generation with FissionVAEs on the CHARM dataset. Best viewed in color.}
\label{fig: mgvae_gen}
\end{figure*}

\subsection{Results and Analysis}
Here we present the following experiments: we first evaluate the overall generative performance of the proposed VAE architectures in both federated and centralized settings, then we explore strategies for encoding the prior distribution $p(z_1)$, and lastly we showcase the use of heterogeneous decoder architectures in our FissionVAEs. For experiments investigating different generation pathways of hierarchical VAEs and the effect of reconstruction losses, please refer to our supplementary material.

\subsubsection{Overall Performance}
The overall performance of the proposed FissionVAE models is summarized in Table \ref{overall}, and generated examples are shown in Fig. \ref{fig: mgvae_gen}. In addition to the FedVAE baseline, a Deep Convolutional GAN (DCGAN) \cite{dcgan} trained via FedGAN \cite{fedGAN} is used for comparison. Since GAN does not directly model the likelihood of data, NLL is not evaluated for FedGAN. Also, FedGAN on CHARM suffers from severe mode collapse, therefore performance evaluation is not available on this dataset. Notably, the performance of all models on the CHARM dataset is less robust compared to the Mixed MNIST dataset. This discrepancy arises because the CHARM dataset, encompassing RGB images from diverse domains, presents a more complex and realistic federated learning scenario. The dataset’s diversity, coupled with a lower local data availability and participation rate among clients, poses greater challenges to federated generative models.

\noindent \textbf{Latent Space Decoupling vs Decoder Branches} As shown in Table \ref{overall}, both latent space decoupling and group-specific decoder branches improve image quality, resulting in lower FID and higher IS. However, decoder branches alone yield greater performance improvements than latent space decoupling, highlighting the negative impact of mixing decoders trained on non-IID data.

Among the configurations, FissionVAE+L shows moderate improvement over the baseline FedVAE by partitioning the latent space according to client groups. This partitioning not only informs the decoder of the expected data characteristics more effectively but also helps to prevent the overlap of data representations from different domains. The effect of latent space decoupling can be also observed in the qualitative results. In Fig. \ref{fig: mgvae_gen}, although FissionVAE+L allows sampling latent codes specific to each group, mixing decoders during aggregation causes artifacts, such as blending features from different domains.

FissionVAE+D greatly alleviates the problem of blended visual features. The FissionVAE+D architecture consists of a unified encoder and decoder branches for individual client groups. The unified encoder acts as a routing mechanism similar to that in a Mixture of Expert (MoE) framework, directing data to the appropriate latent distributions based on its domain. During model aggregation, these decoders among different client groups remain distinct, ensuring that textural features from different domains do not mix, thus producing cleaner and more discernible images as shown in Fig \ref{fig: mgvae_gen}.

FissionVAE+L+D is the architecture that combines both latent space decoupling and decoder branches, Table \ref{overall} shows that there is less gain on the Mixed MNIST dataset while the generative performance is further improved on the CHARM dataset. Enforcing latent space decoupling on FissionVAE+D even lowers the performance on FID. We argue that the main reason for this is the number of client groups. As there are only two groups in Mixed MNIST, decoder branches alone is sufficient for the VAE to differentiate samples from different groups. However, as the number of client groups increases, explicit latent space decoupling provides more direct signal to the VAE about the intra-group difference, allowing the model to better capture the data distribution for different groups. In Fig. \ref{fig: mgvae_gen} it can be observed that images generated by FissionVAE+L+D are sharper than the ones generated by FissionVAE+D.

\noindent \textbf{Hierarchical FissionVAE} As discussed in Section 2, here we consider a hierarchical VAE with two levels of latent variable. In Table \ref{overall}, the architecture FissionVAE+H+L+D performs the best on the CHARM dataset and falls behind its non-hierarchical counterpart on the Mixed MNIST dataset. The hierarchical VAE employs multiple levels of latent representations, which refines the model’s ability to capture and reconstruct complex data distributions more faithfully. The performance degradation on simpler datasets like Mixed MNIST suggests that the hierarchical approach might introduce unnecessary redundancy without proportional gains in performance.

\begin{table}[h]
\centering
\resizebox{\columnwidth}{!}{
\begin{tabular}{cccccc}
\hline
\multicolumn{1}{c|}{\multirow{2}{*}{Model}} & \multicolumn{1}{c|}{\multirow{2}{*}{Prior $p(z_1)$}} & \multicolumn{2}{c|}{Mixed MNIST} & \multicolumn{2}{c}{CHARM} \\ 
\cline{3-6} 
\multicolumn{1}{c|}{} & \multicolumn{1}{c|}{} & \multicolumn{1}{c|}{FID $\downarrow$} & \multicolumn{1}{c|}{IS $\uparrow$} & \multicolumn{1}{c|}{FID $\downarrow$} & IS $\uparrow$ \\ 
\hline
\multirow{5}{*}{FissionVAE+L+D} & identical & \textbf{40.78} & 3.01 & 120.39  & 2.16 \\
 & one-hot & 42.01 & \underline{3.02} & 113.82 & 2.25 \\
 & symmetrical & \underline{41.79} & 2.95 & - & - \\
 & random & 43.26 & 3.00 & \underline{111.77} & \textbf{2.47} \\
 & wave & 42.11 & \textbf{3.04} & \textbf{109.10} & \underline{2.27} \\
 \hline
\multirow{6}{*}{FissionVAE+H+L+D} & identical & 55.91 & 2.96 & 122.16 & \textbf{2.30} \\
 & one-hot & \textbf{53.22} & \underline{2.97} & \underline{121.33} & \underline{2.29} \\
 & symmetrical & 58.21 & \textbf{3.03} & - & -  \\
 & random & 53.99 & 2.94 & 124.91 & 2.23 \\
 & wave & \underline{53.68} & 2.94 & \textbf{118.56} & 2.24 \\
 \cline{2-6}
 & learnable & 47.72 & 2.98 & 107.69 & 2.32 \\ \hline
\end{tabular}
}
\caption{Evaluation of Generation Performance with $z_1$ Priors}
\label{prior_eval}
\end{table}

\begin{table*}[t]
\centering
\begin{tabular}{cccccccccc}
\hline
\multicolumn{1}{c|}{\multirow{2}{*}{\begin{tabular}[c]{@{}c@{}}Decoder Architecture\\ on the FashionMNIST Branch\end{tabular}}} & \multicolumn{3}{c|}{MNIST} & \multicolumn{3}{c|}{FashionMNIST} & \multicolumn{3}{c}{Overall} \\ \cline{2-10} 
\multicolumn{1}{c|}{} & \multicolumn{1}{c|}{FID $\downarrow$}  & \multicolumn{1}{c|}{IS $\uparrow$} & \multicolumn{1}{c|}{NLL $\downarrow$} & \multicolumn{1}{c|}{FID $\downarrow$}  & \multicolumn{1}{c|}{IS $\uparrow$} & \multicolumn{1}{c|}{NLL $\downarrow$} & \multicolumn{1}{c|}{FID $\downarrow$}  & \multicolumn{1}{c|}{IS $\uparrow$} & NLL $\downarrow$\\ 
\hline
Homogeneous & \textbf{46.73} & \textbf{2.41} & \underline{0.38} & \underline{61.81} & \underline{2.92} & \underline{0.61} & \textbf{47.72} & \underline{2.98} & \textbf{0.30} \\
Deeper MLP & 49.54 & \underline{2.38} & \textbf{0.33} & \textbf{60.95} & 2.90 & 0.78 & \underline{48.79} & 2.95 & 0.39 \\
Deeper MLP + Conv & \underline{48.21} & 2.38 & 0.38 & 65.82 & \textbf{2.99} & \textbf{0.60} & 50.16 & \textbf{3.00} & \underline{0.30} \\
\hline
\end{tabular}
\caption{Evaluation of FissionVAE+H+L+D with Heterogeneous Decoder Architectures on the Mixed MNIST}
\label{hetero}
\end{table*}

\subsubsection{Decoupling the Prior of $z_1$} Explicitly decoupling the latent space for different client groups improves the VAE’s ability to generate images that align with the true data distribution (Table \ref{overall}). We explore several priors for the latent distribution, modeled as multivariate Gaussians with customizable means and identity covariance matrices and evaluate them in Table \ref{prior_eval}. Details regarding the formal definition of priors can be found in the supplementary material. 

In non-hierarchical VAEs, $z_1$ represents the sole latent variable, while in hierarchical VAEs, $z_1$ is controlled, with $z_2$ following a standard normal distribution $N(0,1)$. Baseline priors are identical across client groups. Other prior variations include one-hot encoding, symmetrical positive and negative integers, random vectors, wave encodings (with grouped 1s in dimensions corresponding to client groups), and a learnable approach unique to hierarchical VAEs. The learnable approach dynamically aligns priors but sacrifices direct sampling from $p(z_1)$.

Hierarchical FissionVAE often underperforms non-hierarchical variants when predefined priors are used due to increased uncertainty from additional latent layers. However, the learnable approach excels in capturing complex distributions dynamically. In simpler datasets like Mixed MNIST, identical priors suffice, but explicit latent encoding becomes crucial as client group diversity increases, as seen with CHARM.

Among prior definitions, symmetrical priors often lead to divergence on CHARM, as their means may exceed neural network initialization ranges. One-hot and random approaches show comparable results but are less consistent than wave encoding, which clearly distinguishes group priors without out-of-range values.

\subsubsection{Group-level Privacy}
\begin{figure}[h]
\centering
\includegraphics[width=0.9\linewidth]{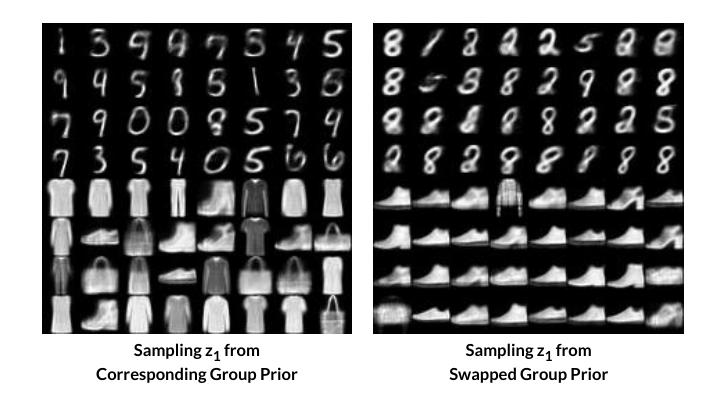}
\caption{In hierarchical FissionVAE, when the prior distribution $p(z_1)$ of the MNIST and FashionMNIST groups are swapped, the generation pathway $q(z_1) \rightarrow q_\phi(z_2|z_1) \rightarrow p_\theta(z_1|z_2) \rightarrow p_\theta(x|z_1)$ leads to sever mode collapse, suggesting potential group-level privacy preserving through protected prior distribution.}
\label{fig: privacy}
\end{figure}

In the presence of hierarchical VAEs, it is possible to incorporate the encoder $q_\phi(z_2|z_1)$ into the generation process, that is, we can first sample the latent code $z_1$ from its prior distribution, then feed it to the subsequent encoder $q_\phi(z_2|z_1)$ and the decoders $p_\theta(z_1|z_2)$ and $p_\theta(x|z_1)$ to obtain the synthesize a generated sample. On the Mixed MNIST dataset, we observe that swapping the prior distributions of the two client groups in the such a generation pathway leads to evident mode collapse, shown in Figure \ref{fig: privacy}. This suggests that the group-level privacy may be preserved by maintaining the confidentiality of prior distributions. This strategy ensures that high-quality samples are generated only when the correct prior distribution is used, while mismatched distributions yield unrecognizable outputs. This phenomenon is more pronounced in both hierarchical and non-hierarchical FissionVAEs on the Mixed MNIST dataset than on the CHARM dataset, likely due to the simpler, more uniform nature of the Mixed MNIST data compared to the diverse and colorful image types in CHARM, which pose greater challenges in satisfying complex latent distribution constraints. Evaluation on other generation pathways are presented in the supplementary material.

\subsubsection{Heterogeneous Decoders in FissionVAE} As discussed in Section 2, the decoupling of decoders for client groups allow for the use of heterogeneous architectures in FissionVAE. The Mixed MNIST dataset, with its relatively simple and grayscale colors, can be generated from both fully connected (MLP) and convolutional layers. In contrast, the more complex and colorful images in the CHARM dataset predominantly require convolutional layers for effective generation.

Table \ref{hetero} details the performance evaluation of various decoder architectures. The term 'homogeneous' refers to identical architectural configurations across all decoder branches, namely a three-layer MLP for each decoder modules. In the `Deeper MLP' configuration, we add two additional fully connected layers to both $p_\theta(z_1|z_2)$ and $p_\theta(x|z_1)$. Meanwhile, we completely replace the decoder $p_\theta(x|z_1)$ from MLP to a series of transpose convolution layers in the `Deeper MLP + Conv' configuration. The results indicate a gradual reduction in overall FID scores as the decoder architecture becomes more heterogeneous. However, the integration of convolutional layers does not improve generation performance over the MLP models, underscoring that while heterogeneous architectures are feasible, they can disrupt the convergence of the VAE due to mismatches in architecture and the model's weight space.

\section{Conclusion}
We presented FissionVAE, a generative model for federated image generation in non-IID data settings. By decoupling the latent space and employing group-specific decoder branches, FissionVAE enhances generation quality while preserving the distinct features of diverse data subsets. Experiments on Mixed MNIST and CHARM datasets demonstrated significant improvements over baseline federated VAE models, with heterogeneous decoder branches and wave-encoded priors proving particularly effective.

Future work includes improving the stability of heterogeneous decoder branches, enabling cross-modality data generation, and developing scalable strategies for handling an increasing number of client groups in real-world federated learning scenarios.

\bibliographystyle{named}
\bibliography{ijcai25}

\clearpage

\section{Supplementary}
In this supplementary material, we provide preliminaries on federated learning (FL) and variational autoencoders (VAEs). We also present the derivation of the the evidence lower bound for our hierarchical VAEs, more qualitative results, discussion on prior distributions and evaluation for different generation pathways. 

\begin{figure*}[ht!]
\centering
\includegraphics[width=1.0\linewidth]{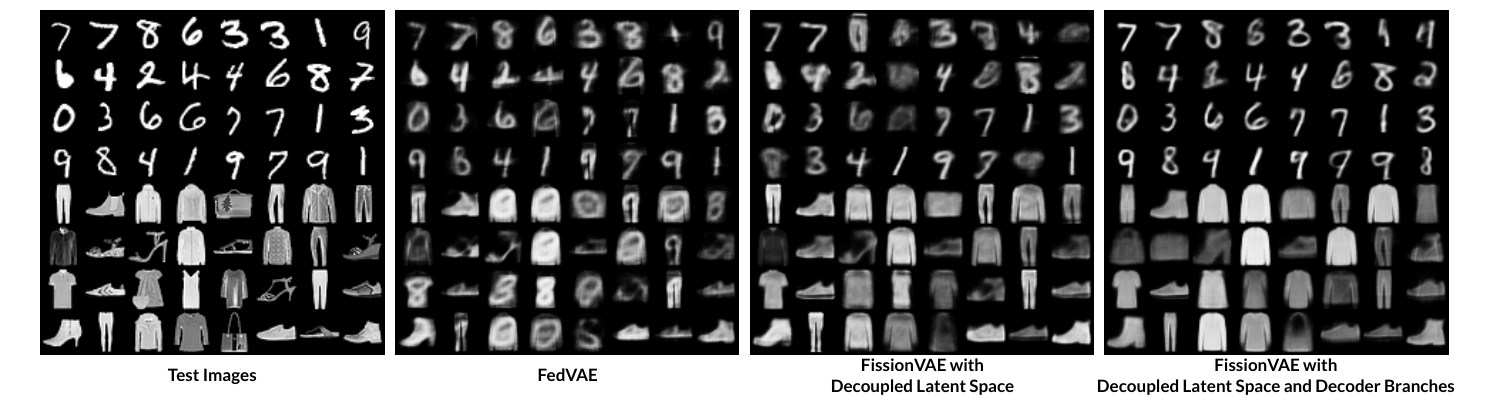}
\caption{Qualitative results of image reconstruction on the Mixed MNIST dataset. As we further decouple the latent space and decoders in FissionVAE, the quality of reconscturcted images gradually improves.}
\label{fig: mnist}
\end{figure*}

\subsection{Preliminaries}
\noindent\textbf{Federated Averaging} - Federated Averaging (FedAvg) \cite{fedavg} is the most commonly used algorithm for federated learning tasks. In FedAvg, the central server first distribute the initialized model to clients, ensuring that all clients starts local training with the identical model parameters. Clients train their models with local data for specified local epochs, then return their models to the central server. The central server aggregates collected client models by taking the weighted average of these models:

\begin{equation}
    \omega_{r} = \sum_{i=1}^{|S|} {\frac{{n_i}}{N_S}\omega_i}
\end{equation}

\noindent where $|S|$ is the number of clients that participate current training round, $n_i$ is the number of training samples on client $i$ while $\omega_i$ is the model parameters for this client, $\omega_r$ is the parameters of the global model, and $N_S$ is the total number of training samples across clients in current round. Once aggregated, the central server distribute the updated model to clients in the next training round. This process is repeated until the global model converges.

\noindent\textbf{Variational Autoencoders} - VAEs \cite{vae-origin} are a class of generative models designed to approximate the probability $p(x)$ of the observed data $x$, which is typically intractable. Central to the optimization of VAEs is the concept of evidence lower bound (ELBO), which provides a tractable objective by bounding the logarithm of $p(x)$ through Jensen's inequality. The ELBO is defined as follows:

\begin{equation}
    \log p(x) = \log \int \frac{p(x, z)q_{\phi}(z|x)}{q_{\phi}(z|x)} dz \geq \underbrace{\mathbb{E}_{q_{\phi}(z|x)} \left[ \log \frac{p(x, z)}{q_{\phi}(z|x)} \right]}_{\rm ELBO}
\end{equation}
where $q_{\phi}(\cdot)$ is the encoder parameterized by $\phi$, and $z$ denotes the latent code derived from the input data $x$. The ELBO can be further decomposed into the following terms by splitting the logarithm,
\begin{equation}
    \rm ELBO = \mathbb{E}_{q_{\phi}(z|x)} [\log p_{\theta}(x|z)] - D_{KL}(q_{\phi}(z|x)||p(z))
\end{equation}
Here, $p_{\theta}(\cdot)$ is the decoder parameterized by $\theta$. The first term of the ELBO represents the expected log-likelihood of the data reconstructed from the latent code $z$. The second term is the Kullback-Leibler divergence between the encoded distribution $q_{\phi}(z|x)$ and a predefined prior distribution $p(z)$. 

During training, the encoder $q_{\phi}$ estimates the latent code $z$ based on the input data $x$, aiming to align this estimate with the prior distribution $p(z)$. The decoder $p_{\theta}$ then attempts reconstruct the original input $x$ from the estimated $z$. The decomposition above indicates that training objective for VAEs is optimizing both the quality of data reconstruction and adherence to the latent space structure. In the generation phase, new data instances are produced by sampling from the prior distribution $p(z)$, which are then transformed solely by the decoder into the data space.

\noindent\textbf{Variational Autoencoders in Federated Learning} - Following the discussion on the ELBO, we note that in a centralized setting, we maximize the expectation of the reconstruction likelihood across all data, as presented in Equation 3. In contrast, federated learning distributes the training samples  $x$ are split into subsets $\{x_1 \cup x_2 \cup ... \cup x_n\}$ across $n$ clients. A common assumption in federated learning is that client data are not identically distributed, yet they are independent. Furthermore, we assume that client data are conditionally independent given the latent codes $z$, such that  $p(x_i|z, x_j) = p(x_i|z), i \neq j$. This leads to the following factorization,

\begin{equation}
\begin{aligned}
    p(x_1, ..., x_n|z) & = p(x_1|z)p(x_2|x_1, z)p(x_3, ..., x_n|z) \\
    &= \Pi_{i=1}^{n}p(x_i|z)
\end{aligned}
\end{equation}

\noindent Therefore, the reconstruction likelihood in the ELBO under the federated assumption can be expressed as:

\begin{equation}
    \log p_{\theta}(x|z) = \Sigma_{i=1}^{n} \log p_\theta(x_i|z)
\end{equation}

\noindent Equation 5 suggests that when training VAEs within the federated learning framework, we can independently maximize the ELBO on each client. Local VAEs are then aggregated through algorithms such as FedAvg to maximize the global ELBO.

\begin{figure*}[ht]
\centering
\includegraphics[width=1.0\linewidth]{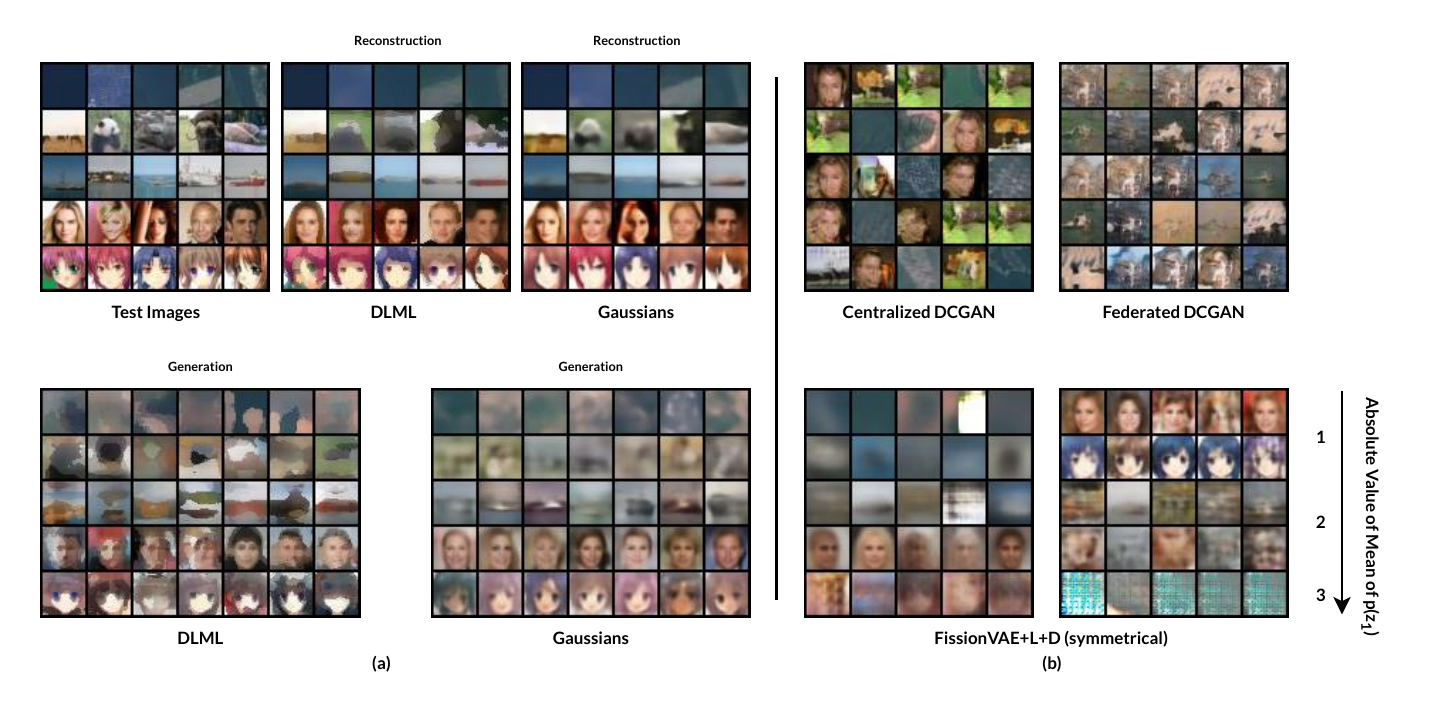}
\caption{Comparison Between Reconstruction Losses and Examples of Mode Collapse. (a) the Discretized Logistic Mixture Likelihood (DLML) loss creates more vibrant colors than the Gaussians, however, it also creates artifacts that resemble watercolor artworks. (b) First Row: FedGAN suffers from severe mode collapse on CHARM in both centralized and federated settings. Second Row: When the ``symmetrical" prior encoding is used on the CHARM dataset, priors that are further away from the origin leads to complete collapse regard. Remote sensing group is assigned a prior mean of +1 on the left, on the right +3 is assigned to this group, resulting in complete collapse.}
\label{fig: loss_compare}
\end{figure*}

\subsection{Derivation of Hierarchical ELBO}
Here we show how the ELBO corresponding to Equation 2 in the main paper is derived. We start by making following assumptions: latent codes are conditionally independent between each other, and the encoding or decoding process is Markovian. Then we can write the encoding and decoding process as,

\begin{gather}
    q(z_1, z_2|x) = q_{\phi}(z_1|x)q_{\phi}(z_2|z_1) \\
    p(x, z_1, z_2) = p(z_2)p_{\theta}(x|z_1)p_{\theta}(z_1|z_2)
\end{gather}

From the log likelihood of the data distribution we have,

\begin{gather}
\begin{aligned}
    \log p(x) & =\log \int p(x, z_{1, 2}) dz_{1, 2} \\ 
    & = \log \int \frac{p(x, z_{1, 2})q(z_{1, 2}|x)}{q(z_{1, 2}|x)}dz_{1, 2} \\ 
    & = \log \mathbb{E}_{q(z_{1, 2}|x)}[\frac{p(x, z_{1, 2})}{q(z_{1, 2}|x)}] \\
    & \ge \mathbb{E}_{q(z_{1, 2}|x)} [\log \frac{p(x, z_{1, 2})}{q(z_{1, 2}|x)}]
\end{aligned}
\end{gather}

Next, we further decompose the lower bound above,

\begin{gather}
    \begin{aligned}
        \text{ELBO} & = \mathbb{E}_{q(z_{1, 2}|x)} [\log \frac{p(x, z_{1, 2})}{q(z_{1, 2}|x)}] \\
        & = \mathbb{E}_{q(z_{1, 2}|x)} [\log \frac{p(z_2)p_{\theta}(x|z_1)p_{\theta}(z_1|z_2)}{q_{\phi}(z_1|x)q_{\phi}(z_2|z_1)}]\\
        & = \mathbb{E}_{q(z_{1, 2}|x)} [\log \frac{p(z_2)p_{\theta}(x|z_1)}{q_{\phi}(z_2|z_1)}]\\
        & \ \ \ \ +\mathbb{E}_{q(z_{1, 2}|x)} [\log \frac{p_{\theta}(z_1|z_2)}{q_{\phi}(z_1|x)}]\\
        & = \mathbb{E}_{q(z_{1, 2}|x)} [\log p_{\theta}(x|z_1)] \\
        & \ \ \ \ + \mathbb{E}_{q(z_{1, 2}|x)} [\log \frac{p(z_2)}{q_{\phi}(z_2|z_1)}] + \mathbb{E}_{q(z_{1, 2}|x)} [\log \frac{p_{\theta}(z_1|z_2)}{q_{\phi}(z_1|x)}] \\
        & = \mathbb{E}_{q_{\phi}(z_1|x)} [\log p_{\theta}(x|z_1)] \\
        & \ \ \ \ - \mathbb{E}_{q_{\phi}(z_{1}|x)} [D_{\text{KL}} (q_{\phi}(z_2|z_1) || p(z_2))] \\
        & \ \ \ \ - \mathbb{E}_{q_{\phi}(z_2|z_1)} [D_{\text{KL}}(q_{\phi}(z_1|x)||p_{\theta}(z_1|z_2)]
    \end{aligned}
\end{gather}

\subsection{Qualitative Results on Mixed MNIST and CHARM}

\begin{figure}[ht]
\centering
\includegraphics[width=1.0\linewidth]{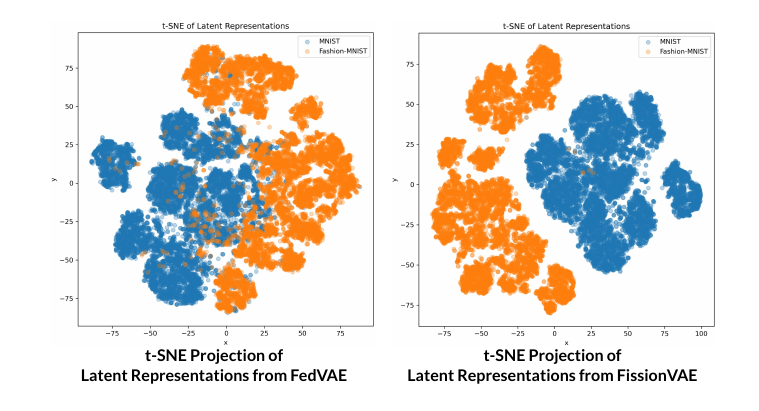}
\caption{t-SNE Projection of Latent Representation on the Mixed MNIST Test Set. Without constraints on decoupling the latent space, the latent space of baseline FedVAE is overlapped and ambiguous. In contrast, FissionVAE learns a more separable latent space.}
\label{fig: tsne}
\end{figure}

Qualitative results of image generation the Mixed MNIST dataset is presented in our main paper. We show the reconstruction on test images in Figure \ref{fig: mnist}. As can be seen, the baseline FedVAE poorly reconstructs the test samples. As we further decouple the latent distributions and decoders, the quality of reconstructed images are improved. We also visualize the learned latent space by projecting representations of test samples through t-SNE \cite{tsne} in Figure \ref{fig: tsne}. Since the baseline FedVAE only considers the divergence between the learned latent distribution and the normal distribution, the learned latent space can be overlapped for different client groups, resulting ambiguity when sampling from the such space during image generation. In FissionVAE we explicitly decouples the latent distribution for different client groups, leading to a more well-structured and separable latent space.

More qualitative results on the CHARM dataset are shown in Figure \ref{fig: mgvae_gen}. As discussed in our main paper, despite that we explicitly decouple the latent distribution for different client groups, as long as their decoders are aggregated uniformly as in FissionVAE+L, the generated images shows noticeable artifacts that blends visual features of distinct image types. The best results are obtained when we decouple both the latent distributions and the decoders, as in FissionVAE+L+D.

\begin{figure*}[ht]
\centering
\includegraphics[width=1.0\linewidth]{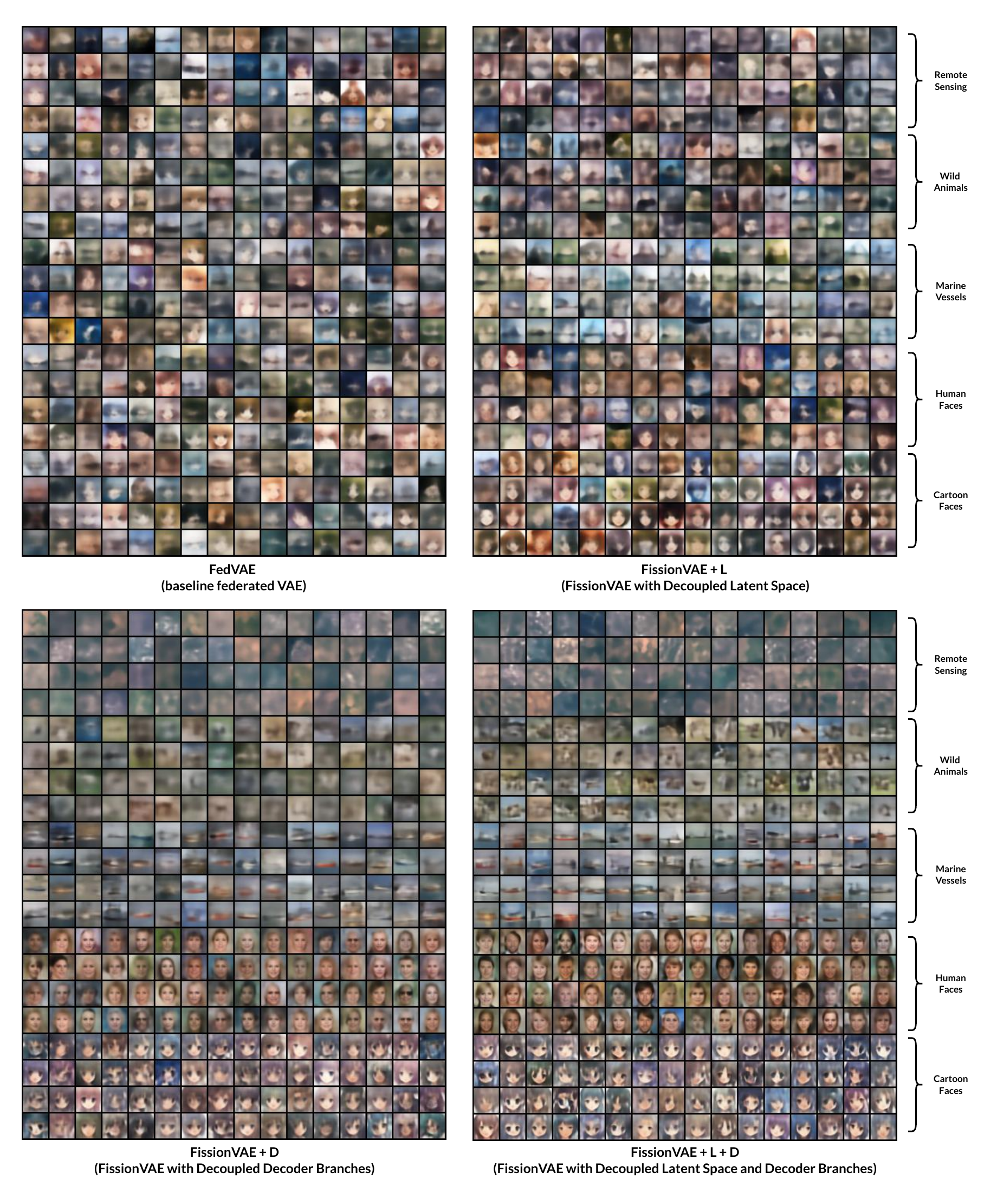}
\caption{Qualitative results of image generation with FissionVAEs on the CHARM dataset. Best viewed in color.}
\label{fig: mgvae_gen}
\end{figure*}

\begin{table*}[ht]
\centering
\begin{tabular}{ccc}
\hline
\multicolumn{1}{c|}{Approach} & \multicolumn{1}{c|}{Definition of Means} & Example of Two Client Groups \\ \hline
identical & $\bm{\mu}^{(j)} = \bm{0}$ & \makecell[c]{[0, 0, 0, 0], [0, 0, 0, 0]} \\
one-hot & $\mu_j^{(j)} = 1$ & [0, 1, 0, 0], [0, 0, 1, 0] \\
symmetrical & $\mu_i^{(j)} = \lceil \frac{j}{2} \rceil$ (j is even) or $ - \lceil \frac{j}{2} \rceil$ (j is odd) & [1, 1, 1, 1], [-1, -1, -1, -1] \\
random & $\mu_i^{(j)} \sim N(0, 1)$ & [0.9, 0.3, 1.6, -0.5], [1.5, -1.8, 0.1, 0.6] \\
wave & $\mu_i^{(j)} = 1, i \in \left[ j \lfloor \frac{dim(z_1)}{k} \rfloor, (j + 1) \lfloor \frac{dim(z_1)}{k} \rfloor \right) $ & [1, 1, 0, 0], [0, 0, 1, 1] \\
learnable & learn from $D_{\text{KL}}(q_{\phi}(z_1|x)||p_{\theta}(z_1|z_2))$ & - \\ \hline
\end{tabular}
\caption{Definition of Means for the Prior Distribution of $z_1$. Elements that are not defined above are 0's by default. $\mu_i^{(j)}$ denotes the $i$-th dimension of the mean for $z_1$ from the client group $j$. $k$ is the total number of client groups.}
\label{prior_def}
\end{table*}

\subsection{Modeling Pixels with Reconstruction Loss}

The reconstruction term in the ELBO quantifies the likelihood of the original input given the estimated distribution from the decoder. L1 or L2 losses, which measure absolute or squared differences, respectively, do not align well with the probabilistic outputs of VAEs. These metrics fail to capture the underlying data distribution effectively, making them less suitable for VAEs that model the complex variability of image pixels.

\begin{table}[H]
\centering
\resizebox{\columnwidth}{!}{
\begin{tabular}{cccc}
\hline
Model & Reconstruction Loss & FID $\downarrow$ & IS $\uparrow$\\
\hline
\multirow{2}{*}{FissionVAE+L+D} & DLML & \textbf{85.12} & \textbf{3.54} \\
 & Gaussian & 109.10 & 2.27 \\
\hline
\multirow{2}{*}{FissionVAE+H+L+D} & DLML & \textbf{82.35} & \textbf{3.91} \\
 & Gaussian & 118.56 & 2.24 \\
 \hline
\end{tabular}
}
\caption{Comparison between Discretized Mixture of Logistics and Gaussians as the reconstruction loss.}
\label{dmol}
\end{table}

\begin{table*}[]
\centering
\begin{tabular}{ccccc}
\hline
\multicolumn{1}{c|}{\multirow{2}{*}{Generation Pathway}} & \multicolumn{2}{c|}{Mixed MNIST} & \multicolumn{2}{c}{CHARM} \\
\cline{2-5} 
\multicolumn{1}{c|}{} & \multicolumn{1}{c|}{FID $\downarrow$} & \multicolumn{1}{c|}{IS $\uparrow$} & \multicolumn{1}{c|}{FID $\downarrow$} & IS $\uparrow$ \\ 
\hline
$q(z_1) \rightarrow q_\phi(z_2|z_1) \rightarrow p_\theta(z_1|z_2) \rightarrow p_\theta(x|z_1)$ & \underline{93.79} & \underline{2.43} & \underline{141.97} & 2.02 \\
$ q(z_1) \rightarrow p_\theta(x|z_1) $ & 124.76 & 2.41 & 144.26 & \underline{2.05} \\
$ p(z_2) \rightarrow p_\theta(z_1|z_2) \rightarrow p_\theta(x|z_1) $ & \textbf{53.68} & \textbf{2.94} & \textbf{118.56} & \textbf{2.24} \\ 
\hline
\end{tabular}
\caption{Evaluation of Image Generation from Different Pathways in FissionVAE+H+L+D}
\label{pathway}
\end{table*}

For grayscale images in the Mixed MNIST dataset, we follow the convention by modeling the reconstruction term using binary cross-entropy between the decoder output and the original input, which captures the presence or absence of pixel values in binary images. For colorful images, modern VAEs often employ the Discretized Logistic Mixture Likelihood (DLML) \cite{dmol}, which models the pixel values more flexibly by considering a mixture of logistic distributions, thus capturing a wider range of color dynamics. In our implementation of FissionVAE, we consider a simpler Gaussian model for each pixel, assuming a shared global variance across all images.

When comparing these two loss function options, we find that while DLML can produce sharper and more visually appealing images, as reflected in higher inception scores, it also introduces stylized artifacts reminiscent of watercolor artworks, characterized by large, irregular color patches. Consequently, we continue to employ the Gaussian formulation in all our implementations due to its simplicity and more consistent performance across varied image types. The qualitative comparison is shown in Figure \ref{fig: loss_compare} and the evaluation results are presented in Table \ref{dmol}.

\subsection{Latent Prior Encodings}
We provide the formal definition of the means of prior distributions discussed in Section 3 - Decoupling the of $z_1$ of the main paper, shown in Table \ref{prior_def}. Notably, although the symmetrical encoding works on Mixed MNIST which has only two client groups, FissionVAEs fail to converge as the number of groups increases. We show failed samples from the symmetrical encoding along with the mode collapse of FedGAN \cite{fedGAN} in Figure \ref{fig: loss_compare}.

\subsection{Generation Pathways in FissionVAE+H} As we incorporate hierachical VAEs, we can sample from either $q(z_1)$ or $p(z_2)$ to produce generated samples. There are two possible generation pathways for $z_1$. The first involves treating random samples from the prior distribution as outputs from the encoder $q_\phi(z_1|x)$, which are then passed to the encoder $q_\phi(z_2|z_1)$ for further processing before the hierarchical decoder outputs the final generated samples. The second pathway treats $z_1$ samples as outputs from the decoder $p_\theta(z_1|z_2)$, directly generating image samples from the final decoder $p_\theta(x|z_1)$. The pathway for $z_2$ follows that of non-hierarchical VAEs, involving only the decoders in the generation process. We list the discussed pathways and their evaluation results in Table \ref{pathway}, and the prior distributions of $z_1$ are defined with the wave approach in Table \ref{prior_def}.

As indicated in Table \ref{pathway}, the conventional generation pathway from $z_2$ consistently outperforms those from $z_1$. Including the encoder in the pathway from $z_1$ allows for mapping random inputs to more meaningful representations for subsequent decoders, enhancing performance compared to the direct generation pathway using $p_\theta(x|z_1)$.

\end{document}